\documentclass[runningheads]{llncs}

\usepackage{eccv}
\usepackage[width=122mm,left=21mm,paperwidth=164mm,height=193mm,top=21mm,paperheight=235mm]{geometry}

\usepackage{eccvabbrv}
\usepackage{graphicx}
\usepackage{booktabs}
\usepackage{array}
\usepackage{float}
\usepackage{placeins}
\usepackage[accsupp]{axessibility}
\usepackage[pagebackref,breaklinks,colorlinks,citecolor=eccvblue]{hyperref}

\newcolumntype{P}[1]{>{\raggedright\arraybackslash}p{#1}}

\begin{document}


\newcommand{\method}{\textsc{FusionRelight}}

\title{\method{}: Relighting Portraits in Real Time via Hybrid Domain Knowledge Fusion}

\titlerunning{\method{}}

\author{Qian Huang, Mayoore Selvarasa Jaiswal,  Zhen Zhong, Rochelle Pereira, and Jianyuan Min}
\institute{NVIDIA}

\authorrunning{Q.~Huang et al.}


\maketitle

\begin{abstract}

Portrait relighting is a low-level vision problem in which physically plausible illumination transfer, identity preservation, and compact real-time inference must be considered together. Iterative diffusion-style methods can synthesize fine detail, but stochastic inference and cost complicate deterministic live video creation; physically grounded relighting preserves identity, but controlled synthetic or light-stage supervision transfers poorly to unconstrained cameras. We present Hybrid Domain Knowledge Fusion (HDKF), a relighting-specific training framework that learns complementary physics, reflectance, and realism priors from synthetic, One-Light-at-A-Time (OLAT), and in-the-wild data, then distills their source-routed supervision into a compact student with clean teacher labels and degraded student inputs. The framework is trained with pixel-aligned RGB, albedo, and normal supervision, providing a simulation substrate for physically grounded low-level relighting. On a held-out OLAT benchmark, HDKF obtains the best MSE, PSNR, and SSIM among evaluated methods while remaining competitive in LPIPS. The distilled model runs in real time at $512\times512$, reaching $11.89$ ms on an RTX 2060 and $1.82$ ms on an RTX 4090.

\end{abstract}

\begin{figure}[t]
    \centering
    \includegraphics[width=0.9\linewidth]{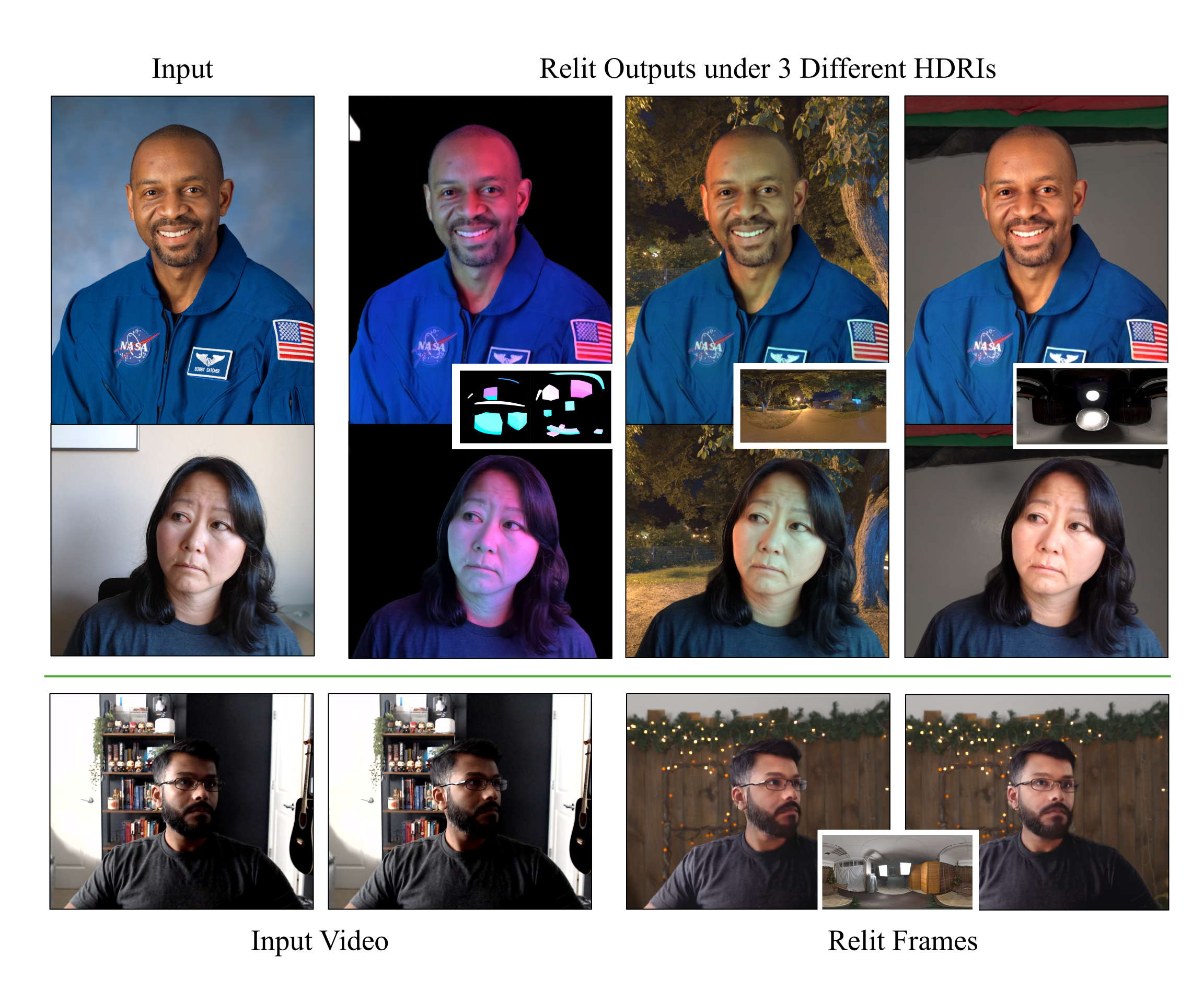}
    \caption{\method{} on images (\emph{top}) and video frames (\emph{bottom}) with target HDR environment maps. The shown outputs change illumination direction and color temperature while retaining subject appearance; target environments are shown as insets.}
    \label{fig:hero-figure}
\end{figure}

\section{Introduction}
\label{sec:intro}

Portrait relighting aims to change the illumination of a person in an image or video according to a target High Dynamic Range Image (HDRI)~\cite{lin2024edgerelight360}. Although the task is often motivated by telepresence, live streaming, and visual effects, its technical core is a low-level vision problem: the output should be equivariant to target illumination while remaining invariant to identity, albedo, fine facial structure, hair, clothing material, and camera artifacts. Supervised in-the-wild pairs with ground-truth intrinsics are not available at scale; synthetic and light-stage data provide physical supervision but differ from real camera captures; and live video requires deterministic, low-latency inference.

Recent relighting systems expose complementary strengths. Many diffusion-style and generative methods produce visually rich results~\cite{zhang2024iclight,he2025unirelight,chaturvedi2025synthlight,mei2025lux,DiffusionRenderer}, but iterative stochastic inference can alter facial details, introduce temporal inconsistency, and remain costly for deterministic live portrait video. Physically grounded methods decompose portraits into reflectance and illumination components~\cite{Pandey21_TotalRelighting,yeh2022lumos,kim2024switchlight,cai2024real}, improving identity preservation and light-transport structure, but models trained primarily on controlled domains often degrade under sensor noise, lens glare, motion blur, ambient leakage, and unfamiliar camera pipelines.

We study deployable portrait relighting as a domain-fusion problem. The key observation is that no single supervision source is reliable for every part of the task: synthetic data provides pixel-aligned intrinsics and controllable rendering, OLAT captures provide measured human reflectance under individual lights, and in-the-wild data provides the camera and identity distribution that the final model must handle. Hybrid Domain Knowledge Fusion (HDKF) couples these sources through three relighting-specific choices: source-domain-routed prior specialization, camera-aware domain adaptation, and asymmetric clean-teacher/corrupted-student distillation into a compact model. To realize this coupling, we choose not to work on new network architectures but to focus on how the models are trained. The contribution is the supervision design for relighting, which prior should label which data, which capture artifacts should be modeled, and how the final student should learn from clean targets while seeing degraded inputs.

In the first stage, we train specialized priors for physical decomposition, OLAT reflectance, and in-the-wild realism. These priors are exposed to forward models for glare, noise, and motion blur, albedo consistency constraints, and 2D Layer Normalization to reduce cross-domain statistical mismatch. In the second stage, frozen priors generate domain-routed pseudo-labels for a compressed student. During distillation, the teacher receives clean inputs while the student receives camera-corrupted inputs. This asymmetry makes the student learn the inverse mapping from degraded observations to clean relighting targets, but only under the assumption that the corruption family preserves the subject structure needed for relighting.

Our contributions in \method{} are:
\begin{itemize}

    \item We formulate deployable portrait relighting as a low-level domain-fusion problem and present HDKF, a relighting-specific training  framework that routes synthetic, OLAT, and in-the-wild data through physics, reflectance, and realism priors before distillation.
    \item We distill the domain-routed priors into a compact student using clean teacher labels and camera-aware corrupted student inputs, obtaining real-time inference at $512\times512$ with measured latency of $11.89$ ms on RTX 2060 and $1.82$ ms on RTX 4090 while preserving quality on a held-out OLAT benchmark.
    \item We construct a large, synthetic relighting dataset with pixel-aligned RGB, alpha, albedo, and normal supervision, and analyze how camera augmentations, albedo consistency, normalization, and multi-teacher distillation affect failure modes.
\end{itemize}

\section{Related Work}
\label{sec:related_work}
\subsection{Portrait Relighting and Physically Grounded Rendering}

Portrait relighting methods often decompose the subject into geometry, reflectance, and illumination before re-rendering under a target environment map. Total Relighting~\cite{Pandey21_TotalRelighting}, LUMOS~\cite{yeh2022lumos}, SwitchLight~\cite{kim2024switchlight}, and 3D-aware real-time relighting~\cite{cai2024real} show the value of explicit physical structure for preserving identity and material appearance. EdgeRelight360~\cite{lin2024edgerelight360} and other efficient relighting systems further emphasize the need for low-latency deployment. The limitation is that physical supervision typically comes from synthetic renders or controlled light-stage captures, so the resulting models can be brittle under real sensor artifacts and unconstrained camera pipelines.

\subsection{Generative Relighting and Low-Level Vision}

Generative relighting methods use diffusion priors for illumination editing across portraits, objects, and video, including IC-Light~\cite{zhang2024iclight}, UniRelight~\cite{he2025unirelight}, Diffusion Renderer~\cite{DiffusionRenderer}, SynthLight~\cite{chaturvedi2025synthlight}, Lux Post Facto~\cite{mei2025lux}, and PI-Light~\cite{liang2026pi}. Many iterative diffusion systems remain slow or stochastic for deterministic live portrait video, where temporal consistency is also required~\cite{Fang25_RelightVid,Corona25_Yesnt}. Our goal is narrower: a deterministic, physically grounded model for identity-preserving, real-time portrait video.

\subsection{Domain Adaptation and Distillation}

Domain adaptation is central to relighting because clean supervision and real deployment inputs differ substantially. Classical image translation and adaptation losses~\cite{zhu2017unpaired,yeh2022lumos} reduce this gap. Camera-artifact training draws on veiling-glare models~\cite{talvala2007veiling}, synthesized flare for learned removal~\cite{wu2021train}, and spatial augmentation tooling~\cite{info11020125}. Knowledge distillation~\cite{hinton2015distilling,gou2021knowledge}, multi-teacher learning~\cite{you2017learning}, and noisy-student training~\cite{xie2020selftraining} have been used to compress or regularize vision models. HDKF combines source-domain teacher routing with camera-corrupted student inputs and clean teacher targets.

\subsection{Low-Level Vision Systems and Assessment}
Portrait relighting has the following concerns even though its output is guided by a target environment map: camera artifacts must be suppressed like restoration problems, shading and specularity must be physically plausible like rendering problems, and the final representation must be compact enough for repeated video inference. Our evaluation therefore combines a physically grounded OLAT benchmark, in-the-wild qualitative detail inspection, component ablations, and latency measurements instead of relying on a single perceptual score.

\subsection{Synthetic Data and Ground-Truth Relighting Evaluation}

Physically grounded low-level vision benefits from renderable data where intrinsics, illumination, and masks are known. Synthetic humans and controlled capture have long supported relighting and reflectance-field evaluation~\cite{debevec2000acquiring,woodham1980photometric,Pandey21_TotalRelighting,yeh2022lumos}. However, a useful deployment dataset must also expose the model to varied identities, HDR illumination, camera properties, and failure modes. Our synthetic corpus provides aligned intrinsics for supervised physical training, while OLAT data supplies measured reflectance for evaluation and real video supplies the unpaired camera distribution needed for adaptation.

\section{Method}
\label{sec:method}

\begin{figure}[t]
    \centering
    \includegraphics[width=\linewidth]{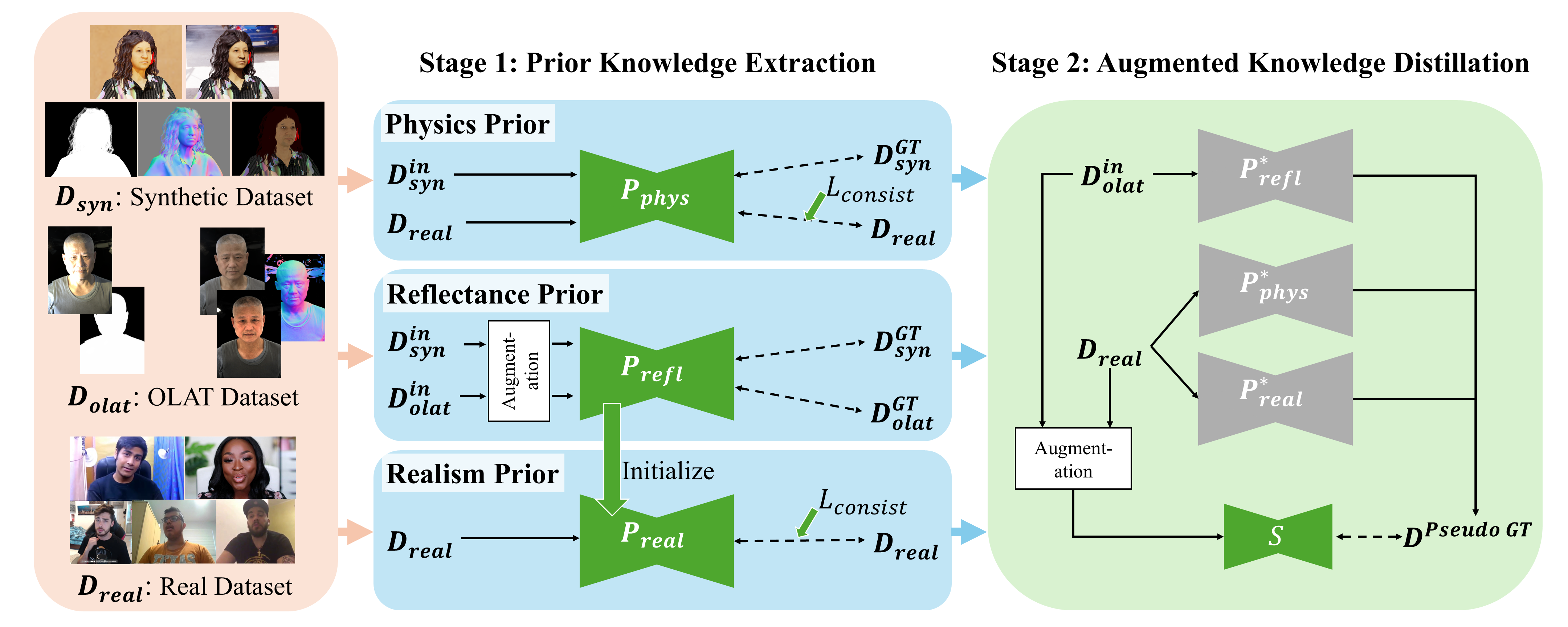}
    \caption{Overview of Hybrid Domain Knowledge Fusion (HDKF). \textbf{Stage~1:} Three specialized prior models are trained on heterogeneous data sources---$\mathcal{P}_{phys}$ on synthetic data with unsupervised real-data adaptation via $\mathcal{L}_{consist}$, $\mathcal{P}_{refl}$ on synthetic and OLAT data with optical augmentations, and $\mathcal{P}_{real}$ initialized from $\mathcal{P}_{refl}$ and fine-tuned on real data. \textbf{Stage~2:} The frozen priors ($\mathcal{P}^{*}$) generate domain-routed pseudo-ground truth for a compact student $\mathcal{S}$. Only the student input is augmented, while the teacher targets remain clean.}
    \label{fig:method}
\end{figure}

\subsection{Deployment-Driven Formulation}

Given a segmented foreground portrait $\mathbf{I}$ and target HDRI $\mathbf{E}$, the relighting model predicts $\hat{\mathbf{R}} = f(\mathbf{I}, \mathbf{E})$. Desired outputs should preserve identity and material appearance from $\mathbf{I}$ while changing shading, cast tone, and specular response according to $\mathbf{E}$. Thus, albedo, facial geometry, hair boundaries, and clothing texture should be invariant to the target environment, whereas shading and highlights should be equivariant to it. Real camera effects such as noise, glare, motion blur, exposure shift, and residual ambient illumination should not be interpreted as relighting cues.

\subsection{Domain-Routed Prior Specialization}

HDKF first trains three prior models, each reliable on a different part of the relighting problem. The physics prior $\mathcal{P}_{phys}$ learns normals, diffuse albedo, and light transport from synthetic data with pixel-aligned intrinsic supervision. The reflectance prior $\mathcal{P}_{refl}$ learns human skin response, specularity, and linear light superposition from synthetic and OLAT data. The realism prior $\mathcal{P}_{real}$ adapts relighting to the long-tail distribution of unconstrained real images and video frames. This decomposition follows the empirical difficulty of balancing heterogeneous supervision in one objective: synthetic labels are physically rich but clean, OLAT captures are measured but controlled, and real data is diverse but unpaired.

Rather than designing a new network, we build all priors on the LUMOS-style decomposition architecture introduced for portrait relighting~\cite{yeh2022lumos}. The foreground is decomposed into surface normal and diffuse albedo maps; the normal indexes precomputed diffuse irradiance and specular environment maps from the target HDRI; and a rendering network predicts per-pixel weights plus an additive residual for the final relit image. We replace standard Batch Normalization with 2D Layer Normalization~\cite{liu2022convnet} in the priors to stabilize feature statistics under heterogeneous synthetic, OLAT, and real-domain batches.

\subsection{Camera-Aware Domain Adaptation}
\label{subsec:domain-adaptation}

To reduce the gap between clean rendered supervision and recorded media, we train with forward models for optical and sensor artifacts. The raw-space augmented image is
\begin{equation}
    \mathbf{I}'_{raw} = (\mathbf{I}_{raw} + \alpha \mathbf{G} + \mathbf{N}_{GAN}) \circledast \mathbf{B},
\end{equation}
    where $\mathbf{G}$ is simulated lens glare, $\alpha$ controls glare strength, $\mathbf{N}_{GAN}$ is learned sensor noise, and $\mathbf{B}$ is a motion-blur kernel. We approximate glare as shift-invariant convolution between synthetic light sources and point-spread functions~\cite{talvala2007veiling,wu2021train}; train NoiseGAN from registered static-camera noise patches~\cite{goodfellow2020generative}; and sample motion kernels with random length and orientation. These corruptions are not intended as a general image-degradation benchmark. They target failure modes that affect portrait relighting because artifacts can be mistaken for facial texture, skin reflectance, or directional illumination.

We also impose albedo consistency on unlabelled real data. For a relighting model $f$, albedo estimator $f_A$, input $\mathbf{I}$, and target HDRIs $\mathbf{E}_1,\mathbf{E}_2$, the environment consistency loss is
\begin{equation}
    \mathcal{L}_{env} =
    \left\| f_A[f(\mathbf{I}, \mathbf{E}_1)] - f_A[f(\mathbf{I}, \mathbf{E}_2)] \right\|_1 .
\end{equation}
Ambient consistency applies brightness or contrast transform $\mathcal{T}$ to the input,
\begin{equation}
    \mathcal{L}_{amb} =
    \left\| f_A[\mathcal{T}(\mathbf{I})] - f_A(\mathbf{I}) \right\|_1 ,
\end{equation}
and the combined loss is
\begin{equation}
    \mathcal{L}_{consist} =
    \lambda_{env}\mathcal{L}_{env} + \lambda_{amb}\mathcal{L}_{amb}.
\end{equation}
Together, these losses make explicit what should not change under relighting: the estimated surface reflectance of the subject.

\subsection{Asymmetric Domain-Routed Distillation}
\label{sec:distillation}

The priors are accurate but too heavy and too domain-specific for deployment. We therefore train a student $\mathcal{S}$ with reduced channel capacity in the normal, albedo, shading, and rendering modules. Routing is fixed by source domain rather than learned confidence: real images use $\mathcal{P}_{real}$, OLAT samples use $\mathcal{P}_{refl}$, and complex-identity synthetic samples use $\mathcal{P}_{phys}$. The selected teacher predicts pseudo-labels for normals, albedos, diffuse and specular weights, light maps, and the final relit output.

The central distillation choice is asymmetric augmentation. Teachers receive the clean input and generate clean pseudo-labels; the student receives an augmented version of the input and is trained to match the clean teacher output. This combines noisy-student augmentation~\cite{xie2020selftraining} with knowledge distillation~\cite{gou2021knowledge}, but our corruptions target relighting deployment artifacts. Albedo remains invariant to target lighting, whereas glare, sensor noise, and motion blur can obscure local evidence. We therefore limit the label-preservation assumption to the sampled corruption ranges in which subject structure remains recoverable. Source-domain routing then trains the student to match clean intrinsic and rendering pseudo-labels under those corruptions.

\section{Datasets}
\label{sec:datasets}

\begin{table}[t]
\centering
\caption{HDKF supervision domains. Synthetic, OLAT, and in-the-wild data respectively provide aligned intrinsics, measured reflectance/ground truth, and camera/identity diversity for adaptation and distillation.}
\label{tab:dataset-roles}
\small
\setlength{\tabcolsep}{3pt}
\begin{tabular}{@{}P{0.16\linewidth}P{0.34\linewidth}P{0.28\linewidth}P{0.16\linewidth}@{}}
\toprule
Source & Scale and supervision & Role in HDKF & Addressed gap \\
\midrule
Synthetic renders & $>640$k rendered frames from 800 avatars and $>2100$ HDRIs; $>340$k 4K samples include pixel-aligned RGB, alpha, diffuse albedo, and normals. & Trains $\mathcal{P}_{phys}$ for intrinsic decomposition and light transport. & Missing paired real intrinsics. \\
OLAT captures & $>200$k relit renders from 320 subjects, 2 expressions, 7 cameras, and 164 LEDs; 32 identities are held out for evaluation. & Trains $\mathcal{P}_{refl}$ and provides ground-truth relighting evaluation. & Skin reflectance and light superposition. \\
In-the-wild data & $>650$k unlabelled real-camera images and video frames with unpaired foregrounds. & Adapts $\mathcal{P}_{real}$ and supplies student distillation diversity. & Camera-domain gap and identity diversity. \\
\bottomrule
\end{tabular}
\end{table}

\subsection{Synthetic Relighting Dataset}

We construct a synthetic corpus using NVIDIA Omniverse and digital human assets. The dataset contains more than 640k synthetic portrait renders from 800 avatars: 300 from a Digital Human Generator API and 500 from RenderPeople. Lighting is sampled from more than 2100 HDRIs, including 1608 real-world, 471 synthetic, and 44 additional varied scene or lighting HDRs. For more than 340k samples, we save pixel-aligned $3840\times2160$ RGB images, alpha mattes, diffuse albedos, and surface normals. To expand diversity, we supplement these renderings with an additional 300k samples rendered from TripleGanger scans at $512 \times 512$.

\subsection{OLAT Data}
\label{sec:olat_data}

The OLAT dataset contains more than 200k renders from 320 subjects captured with two facial expressions, seven frontal cameras, and 164 LEDs. OLAT captures allow target HDR relighting by weighted linear superposition of one-light images~\cite{debevec2000acquiring}, giving pixel-level ground truth for the held-out evaluation in Sec.~\ref{sec:quality}. We reserve 32 identities for testing.

\subsection{Real Data}

The real corpus contains more than 650k unlabelled image and video sequences from two benchmarks, TalkingHead-1kH~\cite{wang2021one} and HD-VILA~\cite{xue2022hdvila}, under diverse cameras, identities, and lighting conditions. Because paired target relighting is unavailable in the wild, these samples are used for unsupervised prior adaptation and teacher-routed student distillation rather than direct supervised evaluation.

\section{Experiments and Results}
\label{sec:experiments}

\subsection{Architecture and Training Details}
\label{sec:training_details}

\subsubsection{Prior Models.}
Our prior models adopt the Phong-based network architecture~\cite{yeh2022lumos}, which decomposes the foreground into surface normal and diffuse
albedo. The predicted normal is used to generate a diffuse light map and multiple specular light maps from pre-convolved environment maps.
A rendering network predicts per-pixel weights to fuse these light maps and estimates an additive residual for the final relit image. The original framework has three training stages: supervised pretraining on a virtual light stage dataset, unsupervised synthetic-to-real albedo adaptation, and temporal refinement with optical flow. These networks use a U-Net architecture. We replace standard Batch Normalization with 2D Layer Normalization and halve the channel capacity and residual-block counts of~\cite{yeh2022lumos}. The normal network operates at $512 \times 512$; the other modules process at $768 \times 768$. We empirically observe that these modifications provide a $4x$ efficiency boost while maintaining comparable visual fidelity.

All prior models use Adam~\cite{kingma2014adam} with learning rate 0.0001 and batch size 8 on 8 NVIDIA A100 GPUs. $\mathcal{P}_{phys}$ is trained from scratch on the synthetic dataset for 100k iterations with the L1, perceptual, and adversarial losses used for surface normals, diffuse albedo, and final RGB as in Total Relighting~\cite{Pandey21_TotalRelighting}. It is then adapted to real data for 50k iterations with $\mathcal{L}_{consist}$ and the unsupervised adaptation objective of LUMOS~\cite{yeh2022lumos}. This unsupervised phase minimizes the composite loss $\mathcal{L} = {\mathcal{L}}_{consist} + \boldsymbol{\lambda}_{adap}^T {\mathcal{L}}_{adap}$, where $\mathcal{L}_{adap} = [\mathcal{L}_{Colat}, \mathcal{L}_{Crela}, \mathcal{L}_{sim}, \mathcal{L}_{Id}, \mathcal{L}_{G}]^T$ enforces lighting consistency, relative lighting consistency, perceptual similarity, and identity preservation.
The weights are empirically set to $\lambda_{env} = 25$ and $\lambda_{amb} = 500$, and $\boldsymbol{\lambda}_{adap} = [10, 50, 10, 1, 1]^T$ corresponding to the respective losses in $\mathcal{L}_{adap}$.

$\mathcal{P}_{refl}$ is trained jointly on our synthetic and OLAT datasets for 100k iterations. Glare augmentation is applied with probability $p=0.5$. The specific augmentation parameters are uniformly sampled within the given range: the PSF standard deviation $\sigma \in [50, 200]$ pixels, shape $\beta \in [1.0, 3.0]$, and glare intensity $\alpha \in [1.0, 3.0]$. Motion blur $k \in [15, 25]$ pixels and $\theta \in [0, 2\pi]$ radians, along with spatial augmentations (resize, rotation, crop) via Albumentations~\cite{info11020125} are applied with $p=0.5$ probability.

$\mathcal{P}_{real}$ is initialized from the $\mathcal{P}_{refl}$ checkpoint and fine-tuned on our real data. This phase utilizes the same unsupervised configuration used in $\mathcal{P}_{phys}$, except that $\lambda_{sim}$ and $\lambda_{Id}$ are doubled to 2 to enforce strict identity fidelity.

\subsubsection{Distilled Student Model.}
\label{sec:training_distillation}
The compressed student network $\mathcal{S}$ is trained for 125k iterations at $768 \times 768$ resolution. Optimization utilizes the Adam optimizer with a batch size $2$ per GPU across 16 NVIDIA A100 GPUs and increased learning rate of 0.0002.

Distillation relies on an explicit, domain-routed batch composition mapping inputs to specific teachers: $75\%$ of each batch ($60\%$ curated images, $15\%$ HD-VILA frames) is supervised by $\mathcal{P}_{real}$; $10\%$ (OLAT data) is supervised by $\mathcal{P}_{refl}$; and the remaining $15\%$ (a residual pool isolating complex-identity subjects) is supervised by $\mathcal{P}_{phys}$.

During asymmetric augmentation, student inputs are degraded with probability $p=0.5$. The distillation objective uses the supervised losses of $\mathcal{P}_{phys}$ plus a differentiable color-histogram loss~\cite{risser2017histogram,afifi2021histogan} for global color and illumination distributions. A temporal residual module~\cite{yeh2022lumos} is then trained on randomly selected adjacent video frames for 50k iterations.

\subsection{Evaluation Setup}

We evaluate HDKF along four axes: OLAT fidelity, in-the-wild visual behavior, efficiency, and component attribution. Quantitative fidelity uses held-out OLAT identities with pixel-aligned targets; in-the-wild evidence is qualitative because paired target relighting is unavailable. Latency evidence combines measured HDKF timings with reported cross-paper context. Metrics are computed on foreground masks: MSE, PSNR, SSIM~\cite{wang2004ssim}, and LPIPS~\cite{zhang2018lpips}.

\textbf{Baselines and scope.}

OLAT baselines are LUMOS~\cite{yeh2022lumos}, DiffusionRenderer~\cite{DiffusionRenderer}, UniRelight~\cite{he2025unirelight}, IC-Light~\cite{zhang2024iclight}, and SwitchLight v3~\cite{kim2024switchlight}. For methods with public implementations, we use the public model/code path available to us, the same target HDRIs, and the same foreground masks for metric computation. SwitchLight v3 is evaluated through its released portal. 3D-Aware Relighting~\cite{cai2024real} operates on tightly cropped faces and does not relight the full upper body, so it is used in qualitative comparisons but excluded from the full-portrait quantitative table. We discuss SynthLight~\cite{chaturvedi2025synthlight} and Lux Post Facto~\cite{mei2025lux} as closely related diffusion relighting systems, but we do not report unverified numbers for methods without comparable outputs in our evaluation setup.

\subsection{Ground-Truth OLAT Fidelity}
\label{sec:quality}

We evaluate on the 32 held-out OLAT identities described in Sec.~\ref{sec:olat_data}. Ground-truth relit images are synthesized by linearly superimposing per-light OLAT frames weighted by target in-the-wild HDR environment maps~\cite{debevec2000acquiring}. This gives pixel-aligned references for all metrics.

\begin{figure}[!htbp]
\centering
\includegraphics[width=0.9\linewidth]{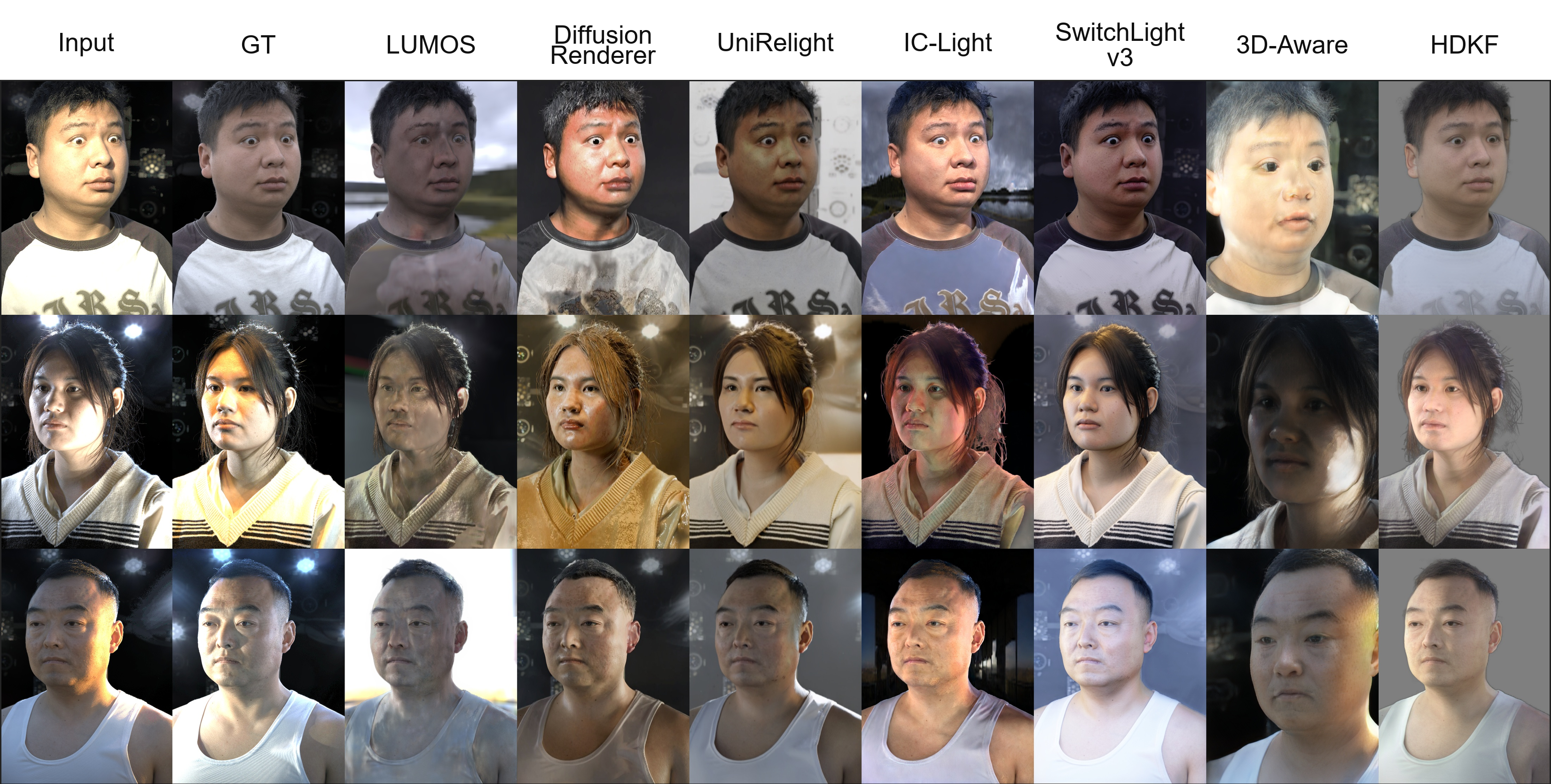}
\caption{Qualitative OLAT comparison: (a)~input, (b)~ground truth, (c)~LUMOS~\cite{yeh2022lumos}, (d)~Diffusion Renderer~\cite{DiffusionRenderer}, (e)~UniRelight~\cite{he2025unirelight}, (f)~IC-Light~\cite{zhang2024iclight}, (g)~SwitchLight v3~\cite{kim2024switchlight}, (h)~3D-Aware Relighting~\cite{cai2024real} (\emph{face crop only}), and (i)~\textbf{Ours}. Table~\ref{tab:quantitative} reports the corresponding full-portrait metrics.}
\label{fig:olat_comp}
\end{figure}

Fig.~\ref{fig:olat_comp} shows qualitative OLAT comparisons. In these examples, HDKF appears closer to the ground-truth brightness, lighting direction, and skin tone than the shown alternatives. Table~\ref{tab:quantitative} shows that HDKF obtains the best MSE, PSNR, and SSIM among evaluated methods. SwitchLight v3 has the best LPIPS; the quantitative conclusion is therefore limited to the three fidelity metrics led by HDKF.

\begin{table}[!htbp]
\centering
\caption{Foreground-masked OLAT results. Best values are \textbf{bold}. HDKF leads MSE/PSNR/SSIM among evaluated full-portrait methods; SwitchLight v3 leads LPIPS, so claims are scoped to structural/physical fidelity.}
\label{tab:quantitative}
\begin{tabular}{lcccc}
\toprule
Method & MSE ($\downarrow$) & PSNR ($\uparrow$) & SSIM ($\uparrow$) & LPIPS ($\downarrow$) \\
\midrule
LUMOS~\cite{yeh2022lumos} & 0.103 & 10.19 & 0.71 & 0.35 \\
DiffusionRenderer~\cite{DiffusionRenderer} & 0.117 & 9.87 & 0.61 & 0.42 \\
UniRelight~\cite{he2025unirelight} & 0.093 & 10.80 & 0.68 & 0.39 \\
IC-Light~\cite{zhang2024iclight} & 0.077 & 11.89 & 0.68 & 0.33 \\
SwitchLight v3~\cite{kim2024switchlight} & 0.045 & 14.61 & 0.74 & \textbf{0.29} \\
\textbf{HDKF} & \textbf{0.037} & \textbf{14.75} & \textbf{0.77} & 0.33 \\
\bottomrule
\end{tabular}
\end{table}

\subsection{Fine-Grained In-the-Wild Visual Analysis}

\begin{figure}[t]
\centering
\includegraphics[width=\linewidth]{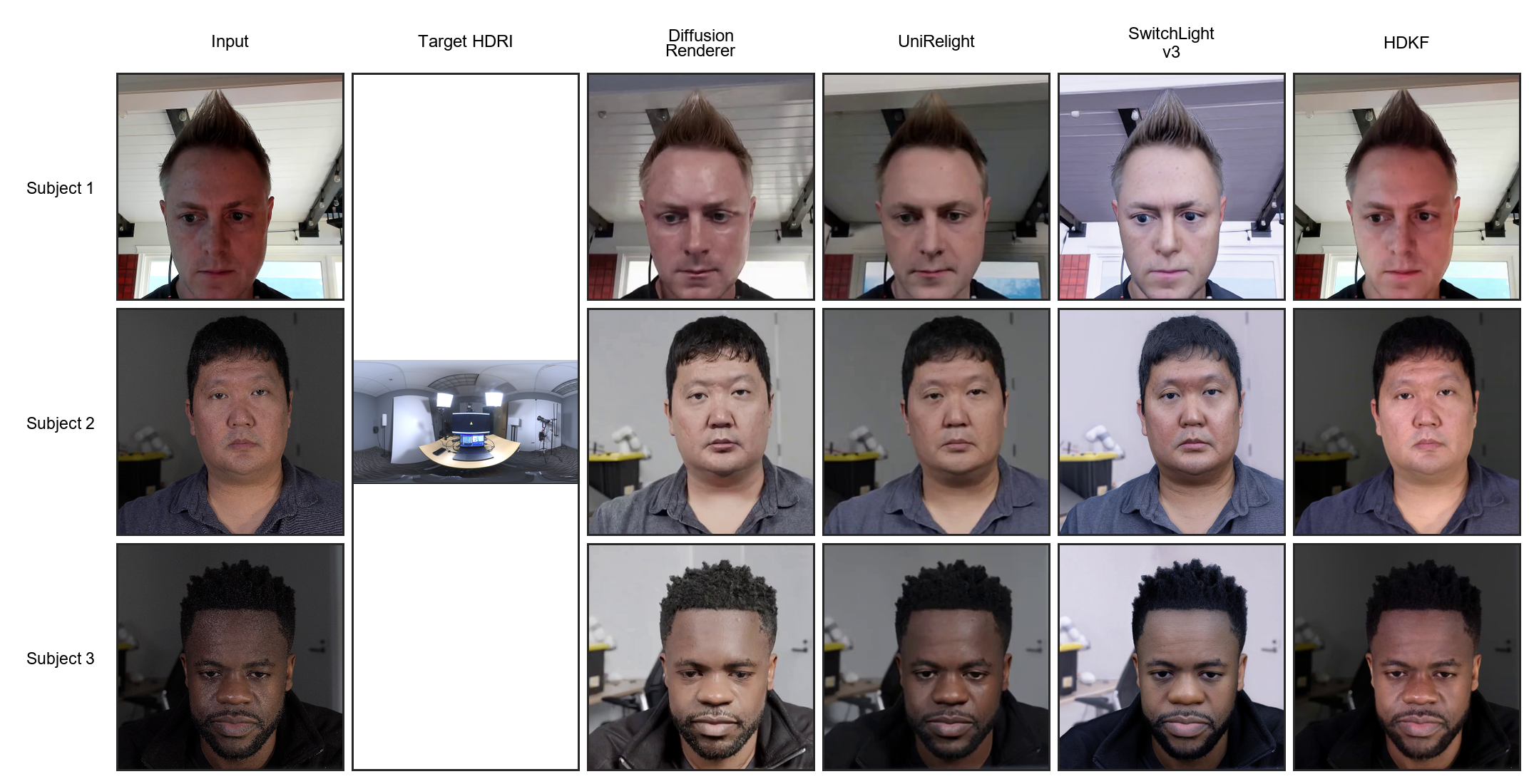}
\caption{Input-anchored crops from in-the-wild relighting. Each row uses one crop window selected on the input and applied unchanged to all outputs; the target HDRI is shown as a thumbnail. The panels support qualitative inspection of detail, boundaries, highlights, and shadow/color transitions.}
\label{fig:visual-crops}
\end{figure}

Real-world relighting has no paired ground truth, so we use qualitative comparisons to inspect identity-sensitive details. Fig.~\ref{fig:visual-crops} uses input-anchored face crops: each crop rectangle is selected once from the input portrait and then applied to the portrait outputs in that row, with only small horizontal registration corrections for baseline columns whose montage alignment is visibly shifted. The crop view makes the main failure modes easier to inspect in these examples: DiffusionRenderer can introduce strong local specularity, UniRelight can under-preserve contrast under some target lighting, and SwitchLight v3 can over-brighten or smooth facial and hair detail. HDKF visually retains subject-specific texture and boundary detail in the shown crops while changing the global illumination direction and color. These comparisons are qualitative; they complement, rather than replace, the OLAT metrics above.

\subsection{Efficiency and Compactness}
\label{sec:latency}

We first report controlled HDKF measurements, then give cross-paper latency context. Unless otherwise stated, HDKF timings are TensorRT-optimized model timings for one foreground frame. We do not treat prior timings as a unified benchmark because publications differ in preprocessing, precision, batching, warmup, and whether timing includes full pipeline overhead.

\begin{table}[t]
\centering
\caption{TensorRT HDKF latency for one $512\times512$ foreground frame; both tested GPUs meet real-time throughput.}
\label{tab:hdkf-latency}
\begin{tabular}{lcc}
\toprule
GPU & Latency (ms) $\downarrow$ & Throughput \\
\midrule
RTX 2060 & 11.89 & $\sim$84 fps \\
RTX 4090 & 1.82 & $>500$ fps \\
\bottomrule
\end{tabular}
\end{table}

\begin{table}[t]
\centering
\caption{Source-aware hardware-matched latency comparison. Prior-method values retain the scope reported in their publications; SwitchLight v3 is the vendor-reported time for VFX-pass generation. HDKF (Our method) values are model-only timings on the listed GPU and resolution. These scopes are not a unified end-to-end benchmark.}
\label{tab:hardware-matched}
\resizebox{\columnwidth}{!}{%
\begin{tabular}{lcccc}
\toprule
Method & GPU & Resolution & Reported latency (ms) $\downarrow$ & Ours (ms) $\downarrow $ \\
\midrule
NVPR~\cite{zhang2021neural} & RTX 3080 & $512\times512$ & 29.2 & \textbf{4.70} \\
LUMOS~\cite{yeh2022lumos} & A6000 & $512\times512$ & 65.0 & \textbf{4.55} \\
Lite2Relight~\cite{rao2024lite2relight} & RTX 3090 & $512\times512$ & 142.9 & \textbf{3.98} \\
SwitchLight v3\footnotemark & RTX 4090 & 2K & 2700.0 & \textbf{11.25} \\
\bottomrule
\end{tabular}%
}
\end{table}
\footnotetext{SwitchLight 3.0 product identity and vendor VFX-pass timing (accessed 2026-07-09): \url{https://beeble.ai/research/switchlight-3-0-is-here}; \url{https://docs.beeble.ai/beeble-studio/setup-requirements}.}

Distillation is central to this deployment profile: on RTX 3080, the prior model runs at $13.53$ ms per frame, while the distilled student runs at $4.70$ ms per frame, a roughly $3\times$ reduction under the same local measurement path.

\subsection{Component Attribution}
\label{sec:ablation}

\begin{figure}[t]
\centering
\includegraphics[width=0.62\linewidth]{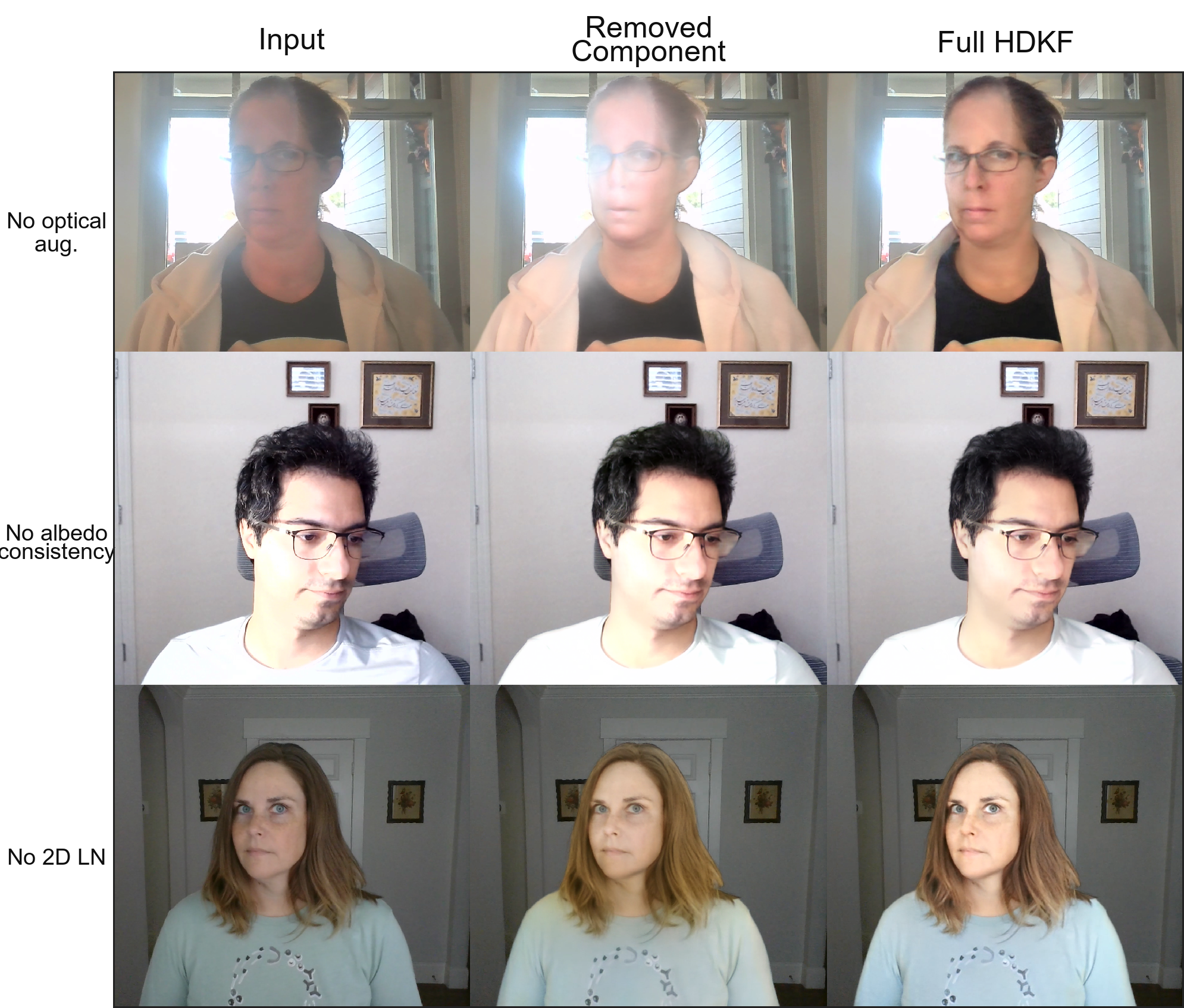}
\caption{Prior-component ablation. Each row removes one training mechanism and compares with full HDKF. The examples show targeted failures: artifacts absorbed as facial texture, ambient leakage without albedo consistency, and weaker contrast/texture without domain-aware normalization.}
\label{fig:ablation-prior}
\end{figure}

\begin{figure}[t]
\centering
\includegraphics[width=0.9\linewidth]{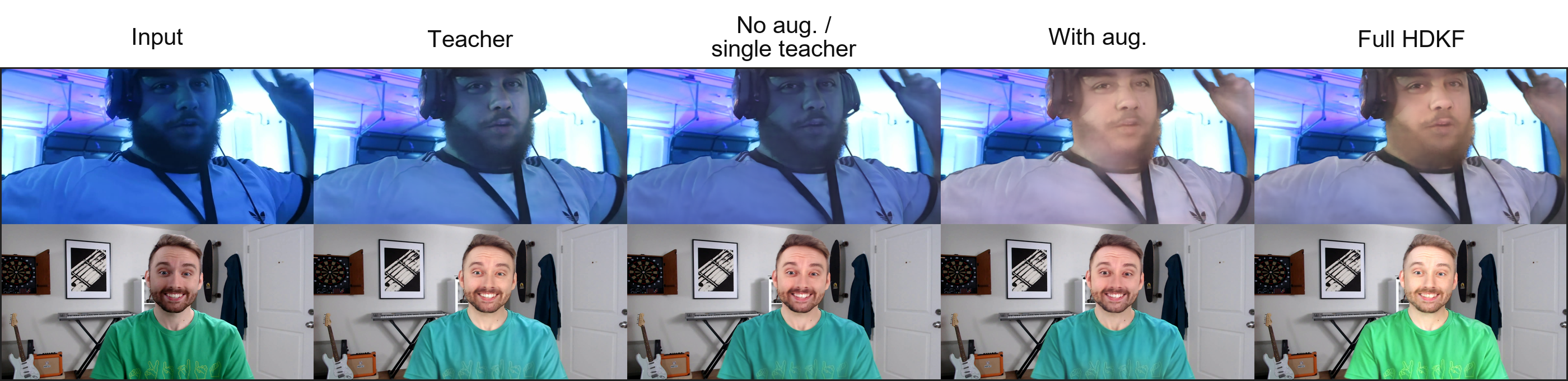}
\caption{Distillation ablation. Columns compare input, teacher, no-augmentation/single-teacher student, augmentation-only variant, and full HDKF. The examples attribute color disentanglement to asymmetric augmentation and identity/skin-tone preservation to domain-routed teachers.}
\label{fig:ablation}
\end{figure}

Fig.~\ref{fig:ablation-prior} and Fig.~\ref{fig:ablation} isolate the main design choices through controlled qualitative ablations. Table~\ref{tab:ablation-interpretation} summarizes the observed failure mode associated with each removed component. The evidence is qualitative, but it is useful for attribution because each ablation targets a distinct relighting failure: confusing optical artifacts with texture, leaking ambient light into the target environment, destabilizing cross-domain statistics, or losing identity through insufficient teacher routing.

\begin{table}[t]
\centering
\caption{Qualitative ablation summary linking each removed mechanism to its observed failure mode and evidence figure.}
\label{tab:ablation-interpretation}
\footnotesize
\begin{tabular}{@{}P{0.30\linewidth}P{0.45\linewidth}P{0.18\linewidth}@{}}
\toprule
Removed component & Observed failure mode & Evidence \\
\midrule
Camera-aware augmentations & Glare and sensor artifacts become facial texture or relighting cues & Fig.~\ref{fig:ablation-prior} \\
Albedo consistency & Ambient light from the input remains in the requested target lighting & Fig.~\ref{fig:ablation-prior} \\
2D Layer Normalization & Cross-domain statistics produce washed-out contrast and weaker texture & Fig.~\ref{fig:ablation-prior} \\
Asymmetric augmentation & Student is less able to disentangle colored illumination from subject appearance & Fig.~\ref{fig:ablation} \\
Multi-teacher routing & Identity and skin-tone preservation degrade relative to domain-routed supervision & Fig.~\ref{fig:ablation} \\
\bottomrule
\end{tabular}
\end{table}

\section{Discussion and Limitations}
\label{sec:discussion}

HDKF is a relighting-specific framework rather than a universal distillation recipe. Its assumptions are most appropriate when camera artifacts preserve the foreground structure needed for relighting and the target domain is covered by the synthetic, OLAT, and real-data mixture. The method is less reliable for severe occlusions, segmentation errors, optics outside the augmentation family, extreme target lighting, and faces or materials underrepresented in the training data. Its Phong-based reflection model may also miss fine-grained specular effects such as rim lighting.

The evaluation also has limits. The OLAT benchmark provides pixel-accurate ground truth but remains a controlled capture setting. In-the-wild and component comparisons are qualitative, and no temporal metric is reported. Cross-source latency values mix our TensorRT model-only timings with pipeline timings from publications and vendor documentation, so they are context rather than a unified benchmark. Public redistribution of the synthetic corpus remains unresolved because some source assets may not be redistributable; no release package or schedule is claimed.

\section{Conclusion}

We presented Hybrid Domain Knowledge Fusion (HDKF), a training framework for real-time, camera-aware portrait relighting. HDKF routes synthetic, OLAT, and in-the-wild data through complementary priors and distills the resulting clean supervision into a compact student trained under modeled camera corruptions. On the reported held-out OLAT benchmark, the student achieves the lowest MSE and highest PSNR and SSIM among the evaluated methods, while attaining real-time TensorRT model-only latency on the tested consumer GPUs. These results suggest that, within an established network architecture, carefully routed supervision can yield greater deployment gains than increased network complexity. The accompanying synthetic relighting corpus and component analysis are intended to support further research on physically grounded, deployment-oriented low-level vision.

\bibliographystyle{splncs04}
\bibliography{main}

\end{document}